\documentclass[11pt]{article}

\usepackage[preprint]{acl}

\usepackage{times}
\usepackage{latexsym}

\usepackage[T1]{fontenc}

\usepackage[utf8]{inputenc}

\usepackage{microtype}

\usepackage{inconsolata}

\usepackage{graphicx}
\usepackage{adjustbox}
\usepackage{booktabs}
\usepackage{diagbox}
\usepackage{subcaption}
\usepackage{amsmath}
\usepackage{multirow}
\usepackage{enumitem}
\usepackage{bm}

\title{Instructions for *ACL Proceedings}

\title{KEO: Knowledge Extraction on OMIn via Knowledge Graphs and RAG for Safety-Critical Aviation Maintenance}
\author{
  Kuangshi Ai\thanks{\ \ Equal contribution.} \\
  \texttt{kai@nd.edu} \\\And
  Jonathan A. Karr Jr\footnotemark[1] \\
  \texttt{jkarr@nd.edu} \\\AND
  Meng Jiang \\
  \texttt{mjiang2@nd.edu} \\\And
  Nitesh V. Chawla \\
  \texttt{nchawla@nd.edu} \\\And
  Chaoli Wang \\
  \texttt{chaoli.wang@nd.edu} \\\AND
  University of Notre Dame
}

\begin{document}
\maketitle
\begin{abstract}
We present Knowledge Extraction on OMIn (KEO), a domain-specific knowledge extraction and reasoning framework with large language models (LLMs) in safety-critical contexts. Using the Operations and Maintenance Intelligence (OMIn) dataset, we construct a QA benchmark spanning global sensemaking and actionable maintenance tasks. KEO builds a structured Knowledge Graph (KG) and integrates it into a retrieval-augmented generation (RAG) pipeline, enabling more coherent, dataset-wide reasoning than traditional text-chunk RAG. We evaluate locally deployable LLMs (Gemma-3, Phi-4, Mistral-Nemo) and employ stronger models (GPT-4o, Llama-3.3) as judges. Experiments show that KEO markedly improves global sensemaking by revealing patterns and system-level insights, while text-chunk RAG remains effective for fine-grained procedural tasks requiring localized retrieval. These findings underscore the promise of KG-augmented LLMs for secure, domain-specific QA and their potential in high-stakes reasoning. The code is available at \href{https://github.com/JonathanKarr33/keo}{https://github.com/JonathanKarr33/keo}.
\end{abstract}

\section{Introduction}
Large Language Models (LLMs) have demonstrated significant potential across various sectors, but they face challenges when applied to highly specialized and safety-critical domains such as aviation maintenance \cite{peykani2025large,zhang2025camb}. These models are often limited by their reliance on parametric knowledge stored within their training data, which can lead to factual inaccuracies, outdated information, and an inability to handle domain-specific, real-world scenarios \cite{hu2024rag}.

A prevailing strategy to address these knowledge deficiencies is Retrieval-Augmented Generation (RAG). However, conventional RAG, which relies on embedding-based retrieval of unstructured text chunks, often results in redundant and fragmented context, hindering performance on tasks requiring complex, multi-hop reasoning or global sensemaking \cite{Ram-TACL2023, Shi-EMNLP2022}. This fragmentation is particularly problematic in safety-critical environments, where the lack of verifiable and transparent context can compromise trust and reproducibility \cite{mealey2025trusted}.

To establish secure and trustworthy AI systems, research has pivoted toward Knowledge Graph (KG)-enhanced RAG. Prior work has shown that integrating KGs, which model knowledge as structured entities and relations \cite{Hogan2021KG, jiang2023evolutionknowledgegraphssurvey}, can effectively mitigate core LLM issues like hallucinations and factual inconsistencies by providing verifiable, domain-specific grounding \cite{wagner-etal-2025-mitigating}. Techniques leveraging relational structures and graph traversal have been explored to enhance reasoning in high-stakes areas like healthcare and risk analysis \cite{zhou2025reflection, Bahr-JIII2025}. Crucially, this structured approach enhances explainability and enables smaller language models to achieve performance comparable to that of larger models, supporting secure deployment in sensitive environments without reliance on external APIs~\cite{perez2024artificial}.

To address these challenges, we introduce KEO (Knowledge Extraction on OMIn), a novel framework that integrates structured knowledge representation with RAG to enhance domain-specific knowledge extraction and reasoning. Our work is grounded in the aviation maintenance domain, leveraging the Operations and Maintenance Intelligence (OMIn) dataset \cite{mealey_2024dataset} to create a new Question Answering (QA) benchmark that evaluates both global sensemaking and fine-grained procedural tasks.

Our core contribution is a methodology that constructs a KG from the OMIn dataset and integrates it into an RAG pipeline. This KG-augmented RAG approach enables more coherent, dataset-wide reasoning, allowing LLMs to uncover complex relationships and system-level insights that are difficult to extract using traditional text-chunk RAG methods. We systematically evaluate several locally deployable LLMs, and our experiments demonstrate that KEO significantly improves performance on complex reasoning tasks, while text-chunk RAG remains effective for localized, procedural QA. These findings highlight the potential of KG-augmented LLMs for secure, accurate, and domain-specific QA, paving the way for their responsible deployment in high-stakes reasoning environments.

\section{Related Work}
\label{sec:relatedwork}

\subsection{OMIn}
The OMIn dataset provides a GS for maintenance data~\cite{mealey_2024dataset} and emphasizes trust and reproducibility due to its focus on sensitive aviation and military information. Given the safety-critical nature of this domain, it is essential that all AI tools for knowledge extraction operate locally without reliance on external APIs~\cite{perez2024artificial}. The dataset consists of 2,748 Aviation Incident records from the Federal Aviation Administration (FAA) \cite{faa_accident_incident_dataset}, each ranging from one to three sentences. Previous work has shown that NLP and LLM performance (e.g., zero-shot F1 scores) on OMIn is low because of its domain-specific content~\cite{mealey2025trusted}. This suggests that the technology readiness levels (TRLs)~\cite{mankins1995technology} of current tools for this data are at a low level (TRLs 1-2). Our goal is to enhance this TRL by applying a KG and RAG approach.

\subsection{Enhancing LLMs with RAG}
RAG enhances LLMs with external information beyond the context window. Retrieval can be integrated by appending documents to prompts~\cite{Ram-TACL2023}, injecting them through cross-attention~\cite{Borgeaud-PMLR2022}, using memory layers for entity representations~\cite{De-arXiv2021, Fevry-EMNLP2020}, or combining outputs with nearest-neighbor token distributions~\cite{Shi-EMNLP2022}. Most approaches still rely on embedding-based text chunks, which often yield redundant context and weaker support for complex reasoning. In contrast, KEO uses KGs to provide structured, dataset-wide context.

\subsection{Knowledge Graphs for RAG}
KGs represent knowledge as entities and relations and provide a flexible way to model complex dependencies~\cite{Hogan2021KG, jiang2023evolutionknowledgegraphssurvey}. For LLMs, they act as external knowledge sources that mitigate knowledge cutoff, factual inconsistency, and hallucination~\cite{hu2024rag}, while improving explainability and domain grounding without retraining~\cite{wagner-etal-2025-mitigating}.

Recent work has explored integrating KGs into RAG to enhance factuality and reasoning. This includes leveraging relational structures~\cite{Procko-AIxSET2024, Gao-arXiv2023, Fan-KDD2024}, supporting KG construction and completion through triple extraction~\cite{Melnyk-EMNLPFindings2022, Trajanoska-arXiv2023}, link prediction~\cite{Yao-ICASSP2025}, and causal discovery~\cite{Zhang-arXiv2024, Ban-arXiv2023}. KG-enhanced RAG methods have investigated subgraph or relational-path prompting~\cite{Baek-NLRSE2023, He-NIPS2024, Sen-NLRSE2023}, grounding outputs~\cite{Ranade-SNAM2023, Kang-arXiv2023}, and retrieval via graph traversal~\cite{Wang-AAAI2024}. These approaches have been applied in domains such as healthcare~\cite{Zhao-WWW2025}, customer service~\cite{Xu-SIGIR2024}, and risk analysis~\cite{Bahr-JIII2025}, where reasoning over structured knowledge is critical.

By incorporating KGs into RAG, LLMs can move beyond simple text-chunk retrieval, enabling more robust reasoning for complex, multi-hop queries that require navigating structured dependencies~\cite{zhou2025reflection}. While GraphRAG~\cite{Edge-arXiv2024} employs graph community summarization, our system KEO instead tailors KG navigation to fragmented aviation maintenance records by embedding entity mentions, expanding with multi-hop neighbors, and applying maximum spanning tree filtering~\cite{zhu-NAACL2025-knowledge}.

\subsection{Using LLMs for Benchmark Construction and Evaluation}
Existing benchmarks for RAG span open-domain~\cite{Yang-EMNLP2018, Chen-AAAI2024, Tang-COLM2024}, medicine~\cite{Xiong-ACLFindings2024}, finance~\cite{Wang-arXiv2024}, and multilingual tasks~\cite{Lyu-TIS2025}. However, these benchmarks largely assess factual retrieval, which suits text-chunk RAG. Recent work shows that LLMs can automatically generate benchmarks involving reasoning and summarization~\cite{Lin-arXiv2023, Yuan-arXiv2024, Wang-arXiv2024}. Inspired by this, KEO creates benchmarks to evaluate global sensemaking by developing questions that require reasoning across records, rather than within single passages, similar to GraphRAG~\cite{Edge-arXiv2024}. Additionally, KEO constructs knowledge-to-action questions to assess knowledge transferability in aviation maintenance.

Recent work in scientific visualization also highlights evaluation-centric agent design~\cite{ai2025evaluation}, benchmark construction~\cite{ai2026scivisagentbench}, literacy assessment~\cite{do2026svlat}, and LLM-assisted interaction~\cite{tang2026texgs, ai2026nli4volvis}, offering complementary perspectives on evaluating domain-focused intelligent systems.

For evaluation, we adopt stronger LLMs as judges, following evidence of high alignment with human ratings~\cite{Zheng-NIPS2023, Gu-arXiv2024, Simret-ACLFindings2025}. Prior work has validated ChatGPT as an effective evaluator~\cite{Wang-arXiv2023}, and RAGAS~\cite{Es-EACL2024} further formalizes LLM-based scoring for RAG. In KEO, LLM-based evaluation spans both fact-based problem–action questions and global sensemaking tasks, combining absolute scoring with pairwise comparisons guided by structured criteria.

\section{Methodology}

\begin{figure*}[ht!]
    \centering
    \includegraphics[width=0.95\textwidth]{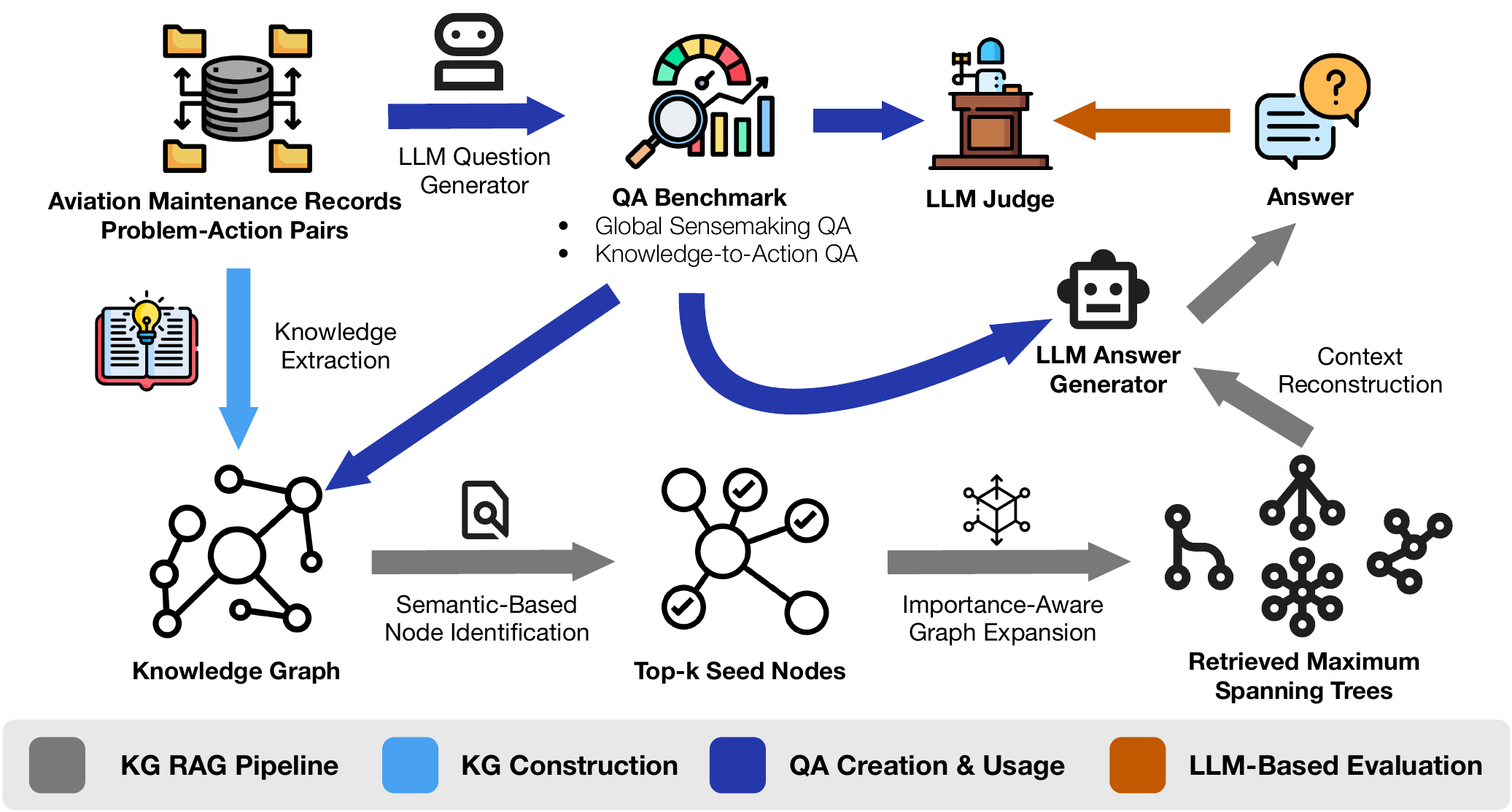}
    \vspace{-0.1in}
    \caption{Overview of the KEO pipeline. Aviation maintenance records and problem–action pairs are first transformed into a QA benchmark covering both global sensemaking and knowledge-to-action questions. In parallel, KEO constructs a structured KG from raw maintenance data. A KG-based RAG workflow then leverages semantic node identification, importance-aware graph expansion, and structured context reconstruction to enhance LLM responses on these safety-critical questions. Finally, an LLM judge evaluates answers through both absolute and comparative scoring with carefully designed metrics.}
    \label{fig:KEO_overview}
\end{figure*}

We present KEO, a domain-specific knowledge extraction and reasoning framework with LLMs in safety-critical settings. As illustrated in Figure~\ref{fig:KEO_overview}, the framework integrates four core components: (1) KG creation from raw maintenance records, (2) a KG-based RAG pipeline for producing safe and reliable answers, (3) automatic creation of an aviation maintenance QA benchmark, and (4) LLM-based evaluation to assess both factual accuracy and higher-level reasoning. Together, these components enable structured sensemaking and actionable decision support in aviation safety applications.

\subsection{LLMs for KG Creation}
Building a KG traditionally depends on human-annotated gold-standard labels for core knowledge extraction tasks such as named entity recognition (NER), coreference resolution (CR), named entity linking (NEL), and relation extraction (RE). However, for the OMIn dataset~\cite{mealey2025trusted}, no gold standard exists for RE (Appendix~\ref{sec:lack_REGS}). To explore the feasibility of automating this initial step, we experimented with both a stronger external model (GPT-4o) and a weaker locally deployable model (Phi-4-mini) in a zero-shot setting. Since knowledge extraction is performed as an offline preprocessing stage, it does not pose major security risks. In contrast, the operational RAG pipeline is carefully designed for secure, local execution, and final answer generation is restricted to lightweight models that can be safely deployed in sensitive environments.

The resulting KG can be represented as a set of weighted triplets:
\begin{equation}
G = \{(h, t, r, w) \mid h \in V,\ t \in V,\ r \in R \}
\end{equation}
where \( h \), \( t \), \( r \), and \( w \) denote the head entity, tail entity, relation, and weight (i.e., frequency of the triplet \( \langle h, r, t \rangle \) in the corpus), respectively. \( V \) is the set of all entity mentions extracted from aviation maintenance records, which is also the set of nodes in KG, and \( R \) is the predefined set of allowed relation types.

Alternatively, this graph can be expressed in standard form as \( G = (V, E) \), where \( V \) is the set of nodes and \( E \subseteq V \times R \times V \) is the set of directed, labeled edges. Each edge \( e = (h, r, t) \in E \) corresponds to a triplet in the original formulation, and can optionally carry a weight \( w(e) \) representing its frequency.

As detailed in Appendix~\ref{sec:kg_creation_prompt}, we construct the KG dynamically with an LLM, iteratively prompting it with the current set of nodes when generating new triplets. This strategy reduces redundancy by minimizing duplicate nodes for the same entity mentions. To evaluate scalability, we progressively build the KG using subsets of the corpus, beginning with 100 records and increasing in steps of 100 up to 500. We evaluate how the performance changes as the size of the underlying corpus grows while ensuring that all nodes can still be accommodated within the LLM’s context window.

\subsection{KG-based RAG Pipeline}
Given the knowledge graph \( G = \{(h, t, r, w)\} \) constructed from the aviation maintenance corpus, the proposed KEO pipeline aims to support global sensemaking and generalizable knowledge-to-action retrieval through a three-step pipeline: semantic-based node identification, importance-aware graph expansion, and KG-based context reconstruction.

\subsubsection{Semantic-Based Node Identification}
In traditional \textit{text-chunk RAG}, the corpus is divided into discrete chunks, denoted as \( C = \{c_1, c_2, \ldots, c_n\} \). Each chunk is embedded into a high-dimensional vector using an embedding model, and the semantic similarity between the user query \( q \) and each chunk \( c \in C \) is computed as:
\begin{equation}
S = \left\{ \frac{\text{emb}(q) \cdot \text{emb}(c)}{\|\text{emb}(q)\| \cdot \|\text{emb}(c)\|} \ \big| \ c \in C \right\}
\end{equation}
where \( \text{emb}(\cdot) \) denotes the embedding function. The top-\( k \) most similar chunks are selected and concatenated into the prompt for LLM inference.

In contrast, KEO performs retrieval at the entity level rather than the chunk level. Specifically, we compute embeddings for each entity mention—that is, each node \( v \in V \) in the knowledge graph—and measure semantic similarity between the user query and the graph nodes:
\begin{equation}
S = \left\{ \frac{\text{emb}(q) \cdot \text{emb}(v)}{\|\text{emb}(q)\| \cdot \|\text{emb}(v)\|} \ \big| \ v \in V \right\}
\end{equation}

A key limitation of \textit{text-chunk RAG} is that retrieved chunks often lack intrinsic structure or semantic cohesion; they may originate from disparate parts of the corpus and present fragmented or redundant information. In contrast, our entity-centric approach retrieves the top-\( k \) most semantically relevant entities $V_k \subset V$ as seed nodes. These seeds serve as entry points for subsequent graph-based expansion and structured context reconstruction, allowing the model to reason over connected and contextually meaningful knowledge.

\subsubsection{Importance-Aware Graph Expansion}
The constructed KG encodes structured relationships among entities, allowing reasoning not only over direct links but also over shared neighbors. For instance, the entities \textit{water in the fuel system} and \textit{fuel tank sumps frozen} may not be directly connected but are both linked to \textit{engine quit}, with edge weights of 8 and 21, respectively. This suggests that while both are plausible causes, \textit{fuel tank sumps frozen} is more frequently observed and potentially more critical. Such patterns underscore the importance of incorporating multi-hop connectivity and edge weights when expanding from the initial seed entities.

Formally, given the initial set of top-\(k\) retrieved seed nodes \( V_k \subseteq V \) and the full graph \( G = (V, E) \), we define the \( m \)-hop expansion subgraph as:
\begin{equation}
G^{(m)} = (V^{(m)}, E^{(m)})
\end{equation}
where \( V^{(m)} \) is the set of nodes reachable from \( V_k \) within \( m \) hops, and \( E^{(m)} \subseteq E \) contains all edges connecting pairs of nodes in \( V^{(m)} \). This expanded subgraph captures both direct and indirect connections, enabling the model to consider not just local matches but also structurally important contextual information.

We later apply importance-aware filtering over \( G^{(m)} \) to prioritize salient and informative substructures before passing them into the context reconstruction module. To enable spanning tree computation, we first transform the directed multi-relational subgraph \( G^{(m)} \) into an undirected weighted graph \( U^{(m)} = (V^{(m)}, \tilde{E}^{(m)}) \). Each undirected edge \( \{u, v\} \in \tilde{E}^{(m)} \) is derived from its directed counterparts in \( E^{(m)} \), where the edge weight is defined as the sum of the weights in both directions, and the relation label is the concatenation of the directed relations (if both exist):

\begin{equation}
\tilde{E}^{(m)} = \big\{\, \big( \{u, v\},\ r',\ w' \big)\}
\end{equation}

where $(u, v, r_{uv}, w_{uv}) \in E^{(m)}$ or $ (v, u, r_{vu}, w_{vu}) \in E^{(m)}$, $w' = w_{uv} + w_{vu}$, and $r' = r_{uv} \Vert r_{vu}$.

Here, \( w_{uv} \) and \( w_{vu} \) are treated as zero if the corresponding directed edge is not present. The operator \( \Vert \) denotes string concatenation. This unified representation allows for computing maximum spanning trees over undirected components while preserving relational semantics.

We then identify all \( l \) connected components in \( U^{(m)} \), denoted as \( \{ H_1, H_2, \dots, H_l \} \), using depth-first search (DFS). For each connected component \( H_i \), we compute its corresponding maximum spanning tree (MST) \( T_i \) using Kruskal’s algorithm:
\begin{equation}
T_i = \text{MST}(H_i), \quad \text{for } i = 1, 2, \dots, l
\end{equation}

These MSTs retain the most structurally significant links within each connected region of the expanded subgraph, ensuring that downstream reasoning is supported by a coherent, importance-aware knowledge structure.

\subsubsection{KG-Based Context Reconstruction}
Given the MSTs ${T_1, T_2, \dots, T_l}$ obtained from the filtered subgraph $U^{(m)}$, we traverse each tree using depth-first search (DFS), starting from the node incident to the edge with the highest weight. For each visited edge, we record its corresponding entities and relation, converting the structured traversal path into textual descriptions. This process yields a KG-based context that retains the original graph structure, serving as input to the downstream LLM.

To further support global sensemaking, we augment the retrieved context with a hierarchical community summarization inspired by GraphRAG~\cite{Edge-arXiv2024}. Specifically, we apply community detection with the Leiden algorithm over the KG to identify densely connected subgraphs and recursively generate summaries for each community level. For leaf communities, we prioritize high-degree nodes and their relations to construct compact summaries. For higher-level communities, we compress summaries of their constituent sub-communities, ensuring a scalable and hierarchical abstraction of the KG. These summaries are concatenated with the graph traversal text to provide the LLM with both local detail and global structure.

\subsection{Automatic QA Benchmark Creation}
\label{sec:QAB}
To construct a robust QA benchmark for aviation safety, we employed state-of-the-art closed-source LLMs, such as GPT-4o, to generate domain-specific questions grounded in aviation maintenance records and problem-action pairs from the OMIn dataset. The benchmark is composed of two major question types: global sensemaking questions and knowledge-to-action questions. The former are designed to require holistic reasoning beyond direct retrieval from the corpus, while the latter evaluate the transferability of knowledge derived from external sources to unseen maintenance problems. To further ensure reliability, we randomly reviewed a subset of the automatically generated questions and confirmed that they met our quality requirements, thereby validating their suitability before incorporating them into the final benchmark.

\subsubsection{Global Sensemaking Questions}
To prevent data leakage and ensure meaningful benchmark creation, we did not expose the question-generation LLMs to the raw maintenance records. Instead, we randomly sampled 500 of the 2,750 OMIn records to construct a knowledge graph, while the remaining 2,250 records were analyzed statistically to extract insights. These insights included failure patterns, component-level distributions, temporal trends, seasonal variations, and aircraft-specific maintenance behaviors.

Using this analytical summary as input, LLMs were prompted to generate high-level, domain-specific questions that require synthesizing information across the dataset—thus assessing global sensemaking. We further classified these questions into three subtypes:
\textbf{(1)} comprehensive questions, which require a holistic understanding of the dataset,
\textbf{(2)} context-specific questions, which focus on patterns within similar maintenance scenarios, and
\textbf{(3)} category-specific questions, which align with predefined analytical dimensions (e.g., failure mode, aircraft type).
The detailed prompts used for each question type are provided in Appendix~\ref{sec:sensemaking_QA_prompt}. Examples of these global sensemaking questions are illustrated in Table~\ref{tab:qa-examples}.

\subsubsection{Knowledge-to-Action Questions}
The MaintNet dataset \cite{akhbardeh2020maintnet} offers targeted problem-action pairs derived from aviation maintenance procedures. To transform these into a QA benchmark, we rephrased each problem as a natural language question using the template: \textit{“What action could be taken when: ...”}, where the corresponding action serves as the gold-standard answer.

While such questions can often be addressed using traditional RAG methods on text chunks, our objective is to evaluate whether knowledge derived from structured maintenance graphs can generalize to unseen cases. To this end, MaintNet records were excluded from the retrievable corpus. Instead, both baseline \textit{text-chunk RAG} and our proposed KEO framework are restricted to retrieving from the OMIn dataset corpus only. This setup ensures that successful answering requires transferring structured maintenance knowledge to new, action-specific scenarios. Illustrative examples of such knowledge-to-action questions and their corresponding gold-standard answers are also shown in Table~\ref{tab:qa-examples}. Details regarding our constrained evaluation setup, including model choice and data limitations, are provided in Appendix \ref{sec:scope}.

\begin{table}[t]
\centering
\vspace{-0.25in}
\caption{Examples of benchmark questions and gold-standard answers of Knowledge-to-Action questions. GSM: Global SenseMaking.}
\label{tab:qa-examples}
\begin{adjustbox}{width=\columnwidth}
\begin{tabular}{lp{6.5cm}}
\toprule
\textbf{Type} & \textbf{Example Question (and Gold-Standard Answer if applicable)} \\
\midrule
GSM (Comprehensive) & What recurring failure modes and seasonal trends emerge across aircraft types, and how might they inform proactive maintenance scheduling? \\
GSM (Context-Specific) & How do patterns of hydraulic leaks in landing gear systems vary across different aircraft classes and operational environments? \\
GSM (Category-Specific) & Which environmental factors are most commonly associated with electrical system failures in narrow-body aircraft? \\
Knowledge-to-Action & \textbf{Q:} What action could be taken when a recurring high-pressure fuel pump vibration is observed during pre-flight inspection?\\
 & \textbf{A:} Replace the high-pressure fuel pump and run a fuel system vibration diagnostic. \\
\bottomrule
\end{tabular}
\end{adjustbox}
\end{table}

\section{Evaluation}

\subsection{Experimental Setup}
We evaluate our method KEO on a benchmark of 133 questions, comprising 83 global sensemaking and 50 knowledge-to-action questions. To construct the KGs, we randomly sample 500 out of 2,748 OMIn records, reserving the remaining 2,248 for question generation. Global sensemaking questions are generated by GPT-4o using insights extracted from these records, while knowledge-to-action questions are created from fixed templates based on problem-action pairs from the MaintNet dataset.

We compare three different methods on this QA benchmark:
\begin{itemize}[leftmargin=10pt, itemsep=0pt, parsep=0pt]
    \item \textbf{Vanilla LLM:} Direct few-shot prompting without any external context from the maintenance corpus.
    \item \textbf{Text-chunk RAG:} Retrieves top-matching text chunks from the maintenance corpus based on embedding similarity.
    \item \textbf{KEO (KG RAG):} Our proposed retrieval method using a KG constructed from OMIn maintenance records.
\end{itemize}

As detailed in Section~\ref{sec:QAB}, the benchmark questions span both global sensemaking and knowledge-to-action tasks, and are generated using GPT-4o, a strong LLM known for aligning well with human preferences~\cite{Achiam-arXiv2023, Shankar-UIST2024, Wei-arXiv2024}. For KG construction, we compare GPT-4o and Phi-4-mini (3.8B, Microsoft) and vary the number of input data points from 100 to 500 to assess scalability.

To test downstream performance on aviation maintenance, we focus on lightweight, locally deployable LLMs for answer generation: Gemma-3-Instruct (27B, Google), Phi-4 (14B, Microsoft), and Mistral-Nemo-Instruct (12B, Mistral AI).

Generated answers are evaluated by two stronger LLM judges, GPT-4o (OpenAI) and Llama-3.3-Instruct (70B, Meta). The judges provide both absolute scores and pairwise comparison scores, following task-specific evaluation metrics in Appendix~\ref{sec:evalmetric} and prompts in Appendix~\ref{sec:evaluator_prompt}.

\begin{figure*}[ht!]
    \centering
    \includegraphics[width=1\linewidth]{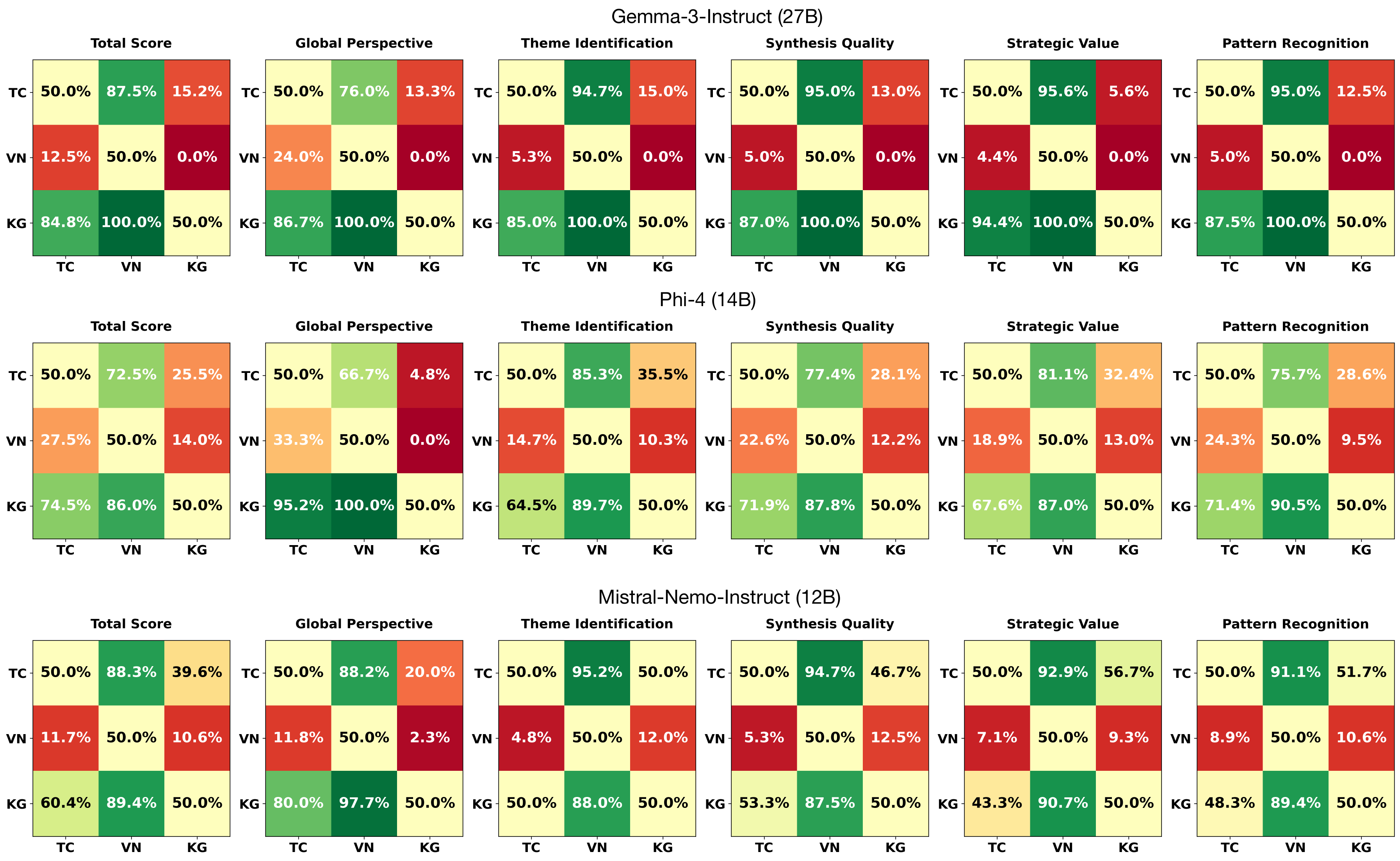}
    \caption{
    Head-to-head win rate matrix of row method over column method (TC: text-chunk RAG, VN: vanilla LLM, KG: our method KEO) for 83 global sensemaking questions, evaluated by GPT-4o. The KG used is generated with GPT-4o from 100 records. Win rates are reported across five dimensions and overall. Green cells indicate a win, red cells indicate a loss. The proposed KEO method consistently outperforms text-chunk RAG when paired with stronger LLMs, but its performance may degrade with weaker backbone models.
    }
    \label{fig:winrate_gpt}
\end{figure*}

\begin{table}[!htbp]
\centering
\caption{Overall evaluation by LLM-based evaluators on 83 global sensemaking questions. The knowledge graph used for KEO is the gold-standard version constructed from 100 records.}
\label{tab:senseQA_overall}
\renewcommand{\arraystretch}{1.3} 
\begin{adjustbox}{max width=\linewidth}
{\large
\begin{tabular}{lcccc}
\toprule
\multirow{2}{*}{\textbf{Model}} & \multirow{2}{*}{\textbf{Evaluator}} & \multicolumn{3}{c}{\textbf{Overall Score (1--5)}} \\
\cmidrule(lr){3-5}
& & \textbf{TC} & \textbf{VN} & \textbf{KG} \\
\midrule
\textbf{gemma-3-it}      & GPT-4o        & $4.12 \pm 0.42$ & $3.70 \pm 0.52$ & \boldmath $4.31 \pm 0.29$ \unboldmath \\
\textbf{phi-4}           & GPT-4o        & $3.90 \pm 0.50$ & $3.68 \pm 0.51$ & \boldmath $4.09 \pm 0.40$ \unboldmath \\
\textbf{mistral-nemo-it} & GPT-4o        & $3.84 \pm 0.61$ & $3.39 \pm 0.60$ & \boldmath $3.87 \pm 0.46$ \unboldmath \\
\midrule
\textbf{gemma-3-it}      & Llama-3.3-it  & $4.47 \pm 0.21$ & $4.29 \pm 0.31$ & \boldmath $4.87 \pm 0.25$ \unboldmath \\
\textbf{phi-4}           & Llama-3.3-it  & $4.33 \pm 0.24$ & $4.32 \pm 0.22$ & \boldmath $4.44 \pm 0.17$ \unboldmath \\
\textbf{mistral-nemo-it} & Llama-3.3-it  & $4.32 \pm 0.25$ & $4.13 \pm 0.42$ & \boldmath $4.39 \pm 0.06$ \unboldmath \\
\bottomrule
\end{tabular}
} 
\end{adjustbox}
\end{table}

\begin{table}[!htbp]
\centering
\caption{Overall evaluation by LLM-based evaluators on 50 knowledge-to-action questions. The knowledge graph used for KEO is the gold-standard version constructed from 100 records.}
\label{tab:actionQA_overall}
\renewcommand{\arraystretch}{1.3} 
\begin{adjustbox}{max width=\linewidth}
{\large 
\begin{tabular}{lcccc}
\toprule
\multirow{2}{*}{\textbf{Model}} & \multirow{2}{*}{\textbf{Evaluator}} & \multicolumn{3}{c}{\textbf{Overall Score (1--5)}} \\
\cmidrule(lr){3-5}
& & \textbf{TC} & \textbf{VN} & \textbf{KG} \\
\midrule
\textbf{gemma-3-it}      & GPT-4o         & $3.84 \pm 0.67$ & $3.75 \pm 0.80$ & \boldmath $3.86 \pm 0.55$ \unboldmath \\
\textbf{phi-4}           & GPT-4o         & \boldmath $4.17 \pm 0.46$ \unboldmath & $4.01 \pm 0.55$ & $3.96 \pm 0.49$ \\
\textbf{mistral-nemo-it} & GPT-4o         & \boldmath $3.72 \pm 0.66$ \unboldmath & $3.70 \pm 0.83$ & $3.68 \pm 0.81$ \\
\midrule
\textbf{gemma-3-it}      & Llama-3.3-it   & \boldmath $3.96 \pm 0.61$ \unboldmath & $3.90 \pm 0.71$ & $3.93 \pm 0.62$ \\
\textbf{phi-4}           & Llama-3.3-it   & \boldmath $4.37 \pm 0.31$ \unboldmath & $4.24 \pm 0.43$ & $4.18 \pm 0.45$ \\
\textbf{mistral-nemo-it} & Llama-3.3-it   & \boldmath $4.00 \pm 0.57$ \unboldmath & $3.78 \pm 0.78$ & $3.80 \pm 0.72$ \\
\bottomrule
\end{tabular}
} 
\end{adjustbox}
\end{table}

\begin{table}[!htbp]
\centering
\caption{ROUGE-based quantitative evaluation on 50 knowledge-to-action questions. We report both ROUGE-1 F1 and ROUGE-L F1 scores. The knowledge graph used for KEO is the gold-standard version constructed from 100 records.}
\label{tab:actionQA_rouge}
\renewcommand{\arraystretch}{1.4}
\begin{adjustbox}{max width=\linewidth}
{\large
\begin{tabular}{lcccc}
\toprule
\textbf{Model} & \textbf{Metric} & \textbf{TC} & \textbf{VN} & \textbf{KG} \\
\midrule
\multirow{2}{*}{\textbf{gemma-3-it}}      
& ROUGE-L & \boldmath $0.276 \pm 0.231$ \unboldmath & $0.273 \pm 0.239$ & $0.272 \pm 0.240$ \\
& ROUGE-1 & \boldmath $0.296 \pm 0.240$ \unboldmath & $0.293 \pm 0.242$ & $0.293 \pm 0.245$ \\
\midrule
\multirow{2}{*}{\textbf{phi-4}}           
& ROUGE-L & $0.093 \pm 0.057$ & $0.100 \pm 0.059$ & \boldmath $0.106 \pm 0.069$ \unboldmath \\
& ROUGE-1 & $0.107 \pm 0.068$ & $0.110 \pm 0.067$ & \boldmath $0.124 \pm 0.077$ \unboldmath \\
\midrule
\multirow{2}{*}{\textbf{mistral-nemo-it}} 
& ROUGE-L & $0.192 \pm 0.195$ & \boldmath $0.271 \pm 0.246$ \unboldmath & $0.268 \pm 0.240$ \\
& ROUGE-1 & $0.213 \pm 0.202$ & \boldmath $0.299 \pm 0.254$ \unboldmath & $0.292 \pm 0.244$ \\
\bottomrule
\end{tabular}
}
\end{adjustbox}
\end{table}

\subsection{Results}
\paragraph{Global Sensemaking Questions.}
As shown in Figure~\ref{fig:winrate_gpt} and Table~\ref{tab:senseQA_overall}, KEO significantly outperforms both vanilla prompting and text-chunk RAG in global sensemaking tasks when evaluated by GPT-4o. The most notable improvements appear in the global perspective criterion and the overall evaluation score across all backbone LLMs. The advantage of KEO is especially pronounced when paired with stronger models such as Gemma-3-Instruct, suggesting that more capable LLMs benefit more from the structured, concise context provided by the KG. In contrast, performance gains are attenuated when KEO is paired with smaller models like Mistral-Nemo-Instruct, likely due to limited reasoning ability to leverage the graph-based context.

\paragraph{Knowledge-to-Action Questions.}
In contrast, as shown in Table~\ref{tab:actionQA_overall}, KEO does not outperform text-chunk RAG on knowledge-to-action questions. Both RAG methods retrieve only from OMIn records to answer questions derived from MaintNet. Because these questions require specific procedural responses, directly retrieving semantically similar records often yields better results than providing abstracted insights from a KG. ROUGE-based evaluations in Table~\ref{tab:actionQA_rouge} show no statistically significant differences, likely due to the short and entity-sparse nature of gold-standard answers, which penalize more comprehensive responses.

\begin{table}[htbp]
\centering
\caption{Evaluation results on 83 global sensemaking questions using GPT-4o as the evaluator. The knowledge graphs employed by the KEO method are constructed by GPT-4o using different numbers of aviation records.}
\label{tab:senseQA_kg-gpt_eval-gpt}
\renewcommand{\arraystretch}{1.4}
\begin{adjustbox}{max width=\columnwidth}
{\Large
\begin{tabular}{lcccc}
\toprule
\textbf{Model} & \textbf{\# of records} & \textbf{TC} & \textbf{VN} & \textbf{KG} \\
\midrule
\multirow{5}{*}{\textbf{gemma-3-it}} 
& 100 & $4.08 \pm 0.55$ & $3.65 \pm 0.68$ & $4.34 \pm 0.24$ \\
& 200 & $4.14 \pm 0.44$ & $3.71 \pm 0.47$ & $4.35 \pm 0.26$ \\
& 300 & $4.11 \pm 0.41$ & \boldmath $3.73 \pm 0.49$ \unboldmath & $4.30 \pm 0.31$ \\
& 400 & \boldmath $4.16 \pm 0.41$ \unboldmath & $3.65 \pm 0.50$ & \fbox{\boldmath $4.38 \pm 0.19$ \unboldmath} \\
& 500 & $4.11 \pm 0.44$ & $3.65 \pm 0.61$ & $4.37 \pm 0.23$ \\
\midrule
\multirow{5}{*}{\textbf{phi-4}} 
& 100 & \boldmath $3.90 \pm 0.49$ \unboldmath & $3.73 \pm 0.51$ & $4.08 \pm 0.39$ \\
& 200 & $3.88 \pm 0.51$ & $3.70 \pm 0.54$ & $4.09 \pm 0.40$ \\
& 300 & $3.87 \pm 0.61$ & \boldmath $3.71 \pm 0.55$ \unboldmath & \fbox{\boldmath $4.11 \pm 0.41$ \unboldmath} \\
& 400 & $3.80 \pm 0.59$ & $3.67 \pm 0.55$ & $4.11 \pm 0.51$ \\
& 500 & $3.89 \pm 0.50$ & $3.69 \pm 0.53$ & $4.09 \pm 0.53$ \\
\midrule
\multirow{5}{*}{\textbf{mistral-nemo-it}} 
& 100 & $3.79 \pm 0.62$ & $3.44 \pm 0.63$ & $3.77 \pm 0.65$ \\
& 200 & $3.80 \pm 0.64$ & \boldmath $3.48 \pm 0.53$ \unboldmath & \fbox{\boldmath $3.90 \pm 0.50$ \unboldmath} \\
& 300 & $3.80 \pm 0.65$ & $3.46 \pm 0.56$ & $3.88 \pm 0.45$ \\
& 400 & \boldmath $3.85 \pm 0.56$ \unboldmath & $3.36 \pm 0.58$ & $3.88 \pm 0.53$ \\
& 500 & $3.76 \pm 0.60$ & $3.45 \pm 0.58$ & $3.86 \pm 0.55$ \\
\bottomrule
\end{tabular}
}
\end{adjustbox}
\end{table}


\paragraph{Ablation Studies.}
We study three factors affecting KEO’s performance: KG size, the LLM used for KG construction, and the choice of evaluator. Tables~\ref{tab:senseQA_kg-gpt_eval-gpt} and \ref{tab:actionQA_kg-gpt_eval-gpt} show GPT-4o’s evaluation across global sensemaking and knowledge-to-action tasks with varying KG sizes. KEO consistently surpasses baselines on global sensemaking, with performance typically peaking when the KG is built from roughly 200--300 records. In contrast, text-chunk RAG remains more competitive for knowledge-to-action tasks, where precise procedural retrieval is crucial. Further analysis in Appendix~\ref{subsec:ablation_KG} shows that KGs produced by weaker models such as Phi-4-mini achieve slightly lower quality than those generated by GPT-4o, but KEO still retains a clear advantage over both vanilla prompting and text-chunk RAG. Evaluator choice also affects relative gains: both GPT-4o and Llama-3.3-70B-Instruct prefer KEO for global sensemaking, while for knowledge-to-action tasks, both evaluators exhibit a mild bias toward text-chunk RAG (Appendix~\ref{subsec:ablation_evaluator}). Overall, these findings underscore the robustness of KEO for dataset-wide reasoning while also revealing the limits of structured abstraction for fine-grained procedural tasks.

\begin{table}[htbp]
\centering
\caption{Evaluation results on 50 knowledge-to-action questions using GPT-4o as the evaluator. The knowledge graphs employed by the KEO method are constructed by GPT-4o using different numbers of aviation records.}
\label{tab:actionQA_kg-gpt_eval-gpt}
\renewcommand{\arraystretch}{1.4}
\begin{adjustbox}{max width=\columnwidth}
{\Large
\begin{tabular}{lcccc}
\toprule
\textbf{Model} & \textbf{\# of records} & \textbf{TC} & \textbf{VN} & \textbf{KG} \\
\midrule
\multirow{5}{*}{\textbf{gemma-3-it}} 
& 100 & $3.87 \pm 0.66$ & \boldmath $3.84 \pm 0.66$ \unboldmath & $3.80 \pm 0.75$ \\
& 200 & \fbox{\boldmath $3.88 \pm 0.65$ \unboldmath} & $3.82 \pm 0.62$ & $3.78 \pm 0.70$ \\
& 300 & $3.84 \pm 0.69$ & $3.83 \pm 0.66$ & $3.78 \pm 0.73$ \\
& 400 & $3.84 \pm 0.67$ & $3.74 \pm 0.74$ & \boldmath $3.80 \pm 0.70$ \unboldmath \\
& 500 & $3.87 \pm 0.67$ & $3.78 \pm 0.68$ & $3.80 \pm 0.77$ \\
\midrule
\multirow{5}{*}{\textbf{phi-4}} 
& 100 & $4.21 \pm 0.41$ & $4.00 \pm 0.54$ & $3.91 \pm 0.55$ \\
& 200 & $4.18 \pm 0.44$ & $4.02 \pm 0.49$ & $3.88 \pm 0.53$ \\
& 300 & $4.18 \pm 0.46$ & $4.03 \pm 0.49$ & \boldmath $3.97 \pm 0.50$ \unboldmath \\
& 400 & \fbox{\boldmath $4.22 \pm 0.39$ \unboldmath} & $4.04 \pm 0.49$ & $3.89 \pm 0.59$ \\
& 500 & $4.17 \pm 0.45$ & \boldmath $4.05 \pm 0.52$ \unboldmath & $3.93 \pm 0.59$ \\
\midrule
\multirow{5}{*}{\textbf{mistral-nemo-it}} 
& 100 & $3.72 \pm 0.67$ & \boldmath $3.72 \pm 0.79$ \unboldmath & $3.64 \pm 0.83$ \\
& 200 & $3.71 \pm 0.68$ & $3.71 \pm 0.81$ & $3.70 \pm 0.82$ \\
& 300 & $3.68 \pm 0.70$ & $3.69 \pm 0.80$ & $3.70 \pm 0.83$ \\
& 400 & \fbox{\boldmath $3.74 \pm 0.71$ \unboldmath} & $3.64 \pm 0.90$ & $3.63 \pm 0.82$ \\
& 500 & \fbox{\boldmath $3.74 \pm 0.71$ \unboldmath} & $3.70 \pm 0.76$ & \boldmath $3.73 \pm 0.85$ \unboldmath \\
\bottomrule
\end{tabular}
}
\end{adjustbox}
\end{table}

\section{Conclusion}
We introduced KEO, a framework integrating KGs into RAG for safety-critical QA. To evaluate its effectiveness, we constructed a benchmark covering both global sensemaking and knowledge-to-action tasks from the OMIn dataset. Experiments with locally deployable LLMs, judged by stronger models, show that KEO substantially improves dataset-wide reasoning and pattern discovery, while text-chunk RAG remains better suited for localized procedural actions. Our results indicate that structured knowledge is crucial for scaling LLMs to high-stakes domains, and that different retrieval paradigms may complement each other depending on task demands.


\newpage
\section*{Limitations}

While our evaluation of KEO focused on the aviation maintenance domain, the framework is transferable to other safety-critical areas such as healthcare, power systems, and defense, where structured reasoning with LLMs can provide actionable insights. Future work should investigate domain adaptation strategies, scaling to larger corpora, and integration with multimodal data sources such as schematics and sensor logs. Due to the scope of work, we limited our study to locally deployable LLMs and had limited public data to use, as noted in Appendix \ref{sec:scope}.

Another limitation lies in evaluation. Although we report quantitative metrics such as ROUGE F1 score, these alone are insufficient to fully capture reasoning quality. To address this, we also adopted LLMs as judges with diverse instruct-tuned models, specialized prompting strategies, and pairwise comparison, which have been shown to improve the robustness of evaluation~\cite{Shankar-UIST2024,Wei-arXiv2024}. However, recent work highlights that LLM-as-a-Judge remains neither fully valid nor reliable: judgments can be highly sensitive to the choice of model, prompt template, and even evaluation order~\cite{chehbouni2025neither}. Moreover, adversarial studies reveal that LLM judges are vulnerable to prompt injection and manipulation~\cite{li2025llmnotjudge}, raising concerns about their robustness in practical settings. These limitations suggest that human-in-the-loop evaluation is essential to complement automated judging and ensure reliable assessment of KEO in safety-critical domains.

Finally, while we relied on locally deployed LLMs to avoid dependency on external APIs, systematic security and robustness checks are needed to mitigate risks such as adversarial prompts, data leakage, and misuse in high-stakes environments.

\section*{Ethical Considerations}
This work, operating in a safety-critical domain, prioritizes ethical deployment by designing the RAG pipeline exclusively for secure, local deployment using smaller, government-approved LLMs to mitigate risks associated with external APIs, data leakage, and potential misuse. We acknowledge that the LLM-constructed KG relies on the sensitive OMIn corpus, which has been processed under strict access controls and de-identification procedures to protect proprietary and personal information. The use of the KG is specifically intended to mitigate the LLM's risk of hallucination and improve factual grounding. However, we emphasize that the system serves only as a decision-support tool, necessitating human validation for all safety-critical applications.


\bibliography{custom}

\newpage
\appendix
\section{Additional Results}
\label{sec:additional_results}
\subsection{Ablation on KG Size and KG Generation LLM}
\label{subsec:ablation_KG}
We assess the impact of KG construction parameters by varying the number of records (from 100 to 500) and the LLM used (GPT-4o vs. Phi-4-mini). Tables~\ref{tab:senseQA_kg-gpt_eval-gpt} and \ref{tab:senseQA_kg-phi_eval-gpt} report GPT-4o evaluations on global sensemaking questions. KEO consistently outperforms baselines, with optimal performance typically achieved when the KG is constructed from 200 to 300 records. While KGs built by Phi-4-mini result in slightly lower scores, KEO still surpasses vanilla and text-chunk RAG methods, highlighting the robustness of our dynamic prompting strategy for high-quality KG construction.

On knowledge-to-action questions (Tables~\ref{tab:actionQA_kg-gpt_eval-gpt} and \ref{tab:actionQA_kg-phi_eval-gpt}), varying KG size and generation LLM does not significantly alter the performance trend—text-chunk RAG remains more effective. This reinforces the observation that specific action-oriented tasks favor localized retrieval over structured abstraction.

\subsection{Ablation on LLM Evaluators}
\label{subsec:ablation_evaluator}
\begin{figure*}[ht!]
    \centering
    \includegraphics[width=1\linewidth]{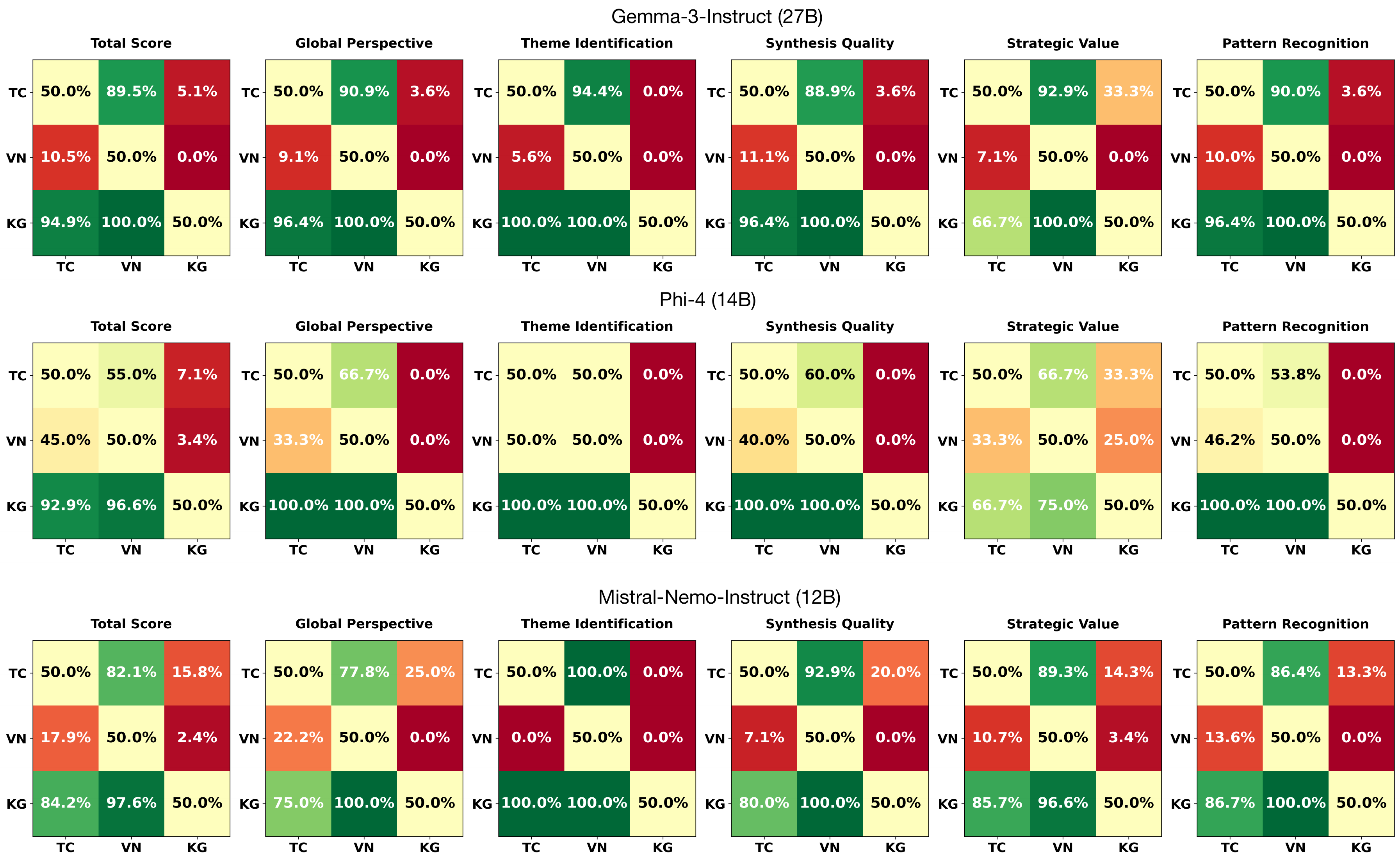}
    \caption{
    Head-to-head win rate matrix of row method over column method (TC: text-chunk RAG, VN: vanilla LLM, KG: our method KEO) on the same 83 global sensemaking questions, evaluated by Llama-3.3-70B-Instruct. Compared to GPT-4o evaluation (Figure~\ref{fig:winrate_gpt}), Llama shows a higher preference for answers generated using RAG-based methods. Nonetheless, the same trend persists: stronger LLMs tend to amplify the advantage of the KEO approach.
    }
    \label{fig:winrate_llama}
\end{figure*}

To evaluate robustness against evaluator bias, we conduct comparative assessments using both GPT-4o and Llama-3.3-70B-Instruct. As shown in Figures~\ref{fig:winrate_gpt} and \ref{fig:winrate_llama}, both evaluators prefer KEO over baselines in global sensemaking tasks, with Llama-3.3 showing even stronger preference for KEO. The trend of stronger LLMs amplifying the performance benefits of KEO remains consistent. Tables~\ref{tab:senseQA_kg-gpt_eval-llama} and \ref{tab:senseQA_kg-phi_eval-llama} confirm this, with optimal KG size again ranging between 200 and 300 records.

For knowledge-to-action questions, Llama evaluations (Tables~\ref{tab:actionQA_kg-gpt_eval-llama} and \ref{tab:actionQA_kg-phi_eval-llama}) mirror those of GPT-4o, favoring text-chunk RAG slightly over KEO. This consistency across evaluators increases confidence in the observed performance tradeoffs between tasks that benefit from structured reasoning and those that require precise, localized retrieval.

\begin{table*}[htbp]
\centering
\caption{Evaluation results on 83 global sensemaking questions using GPT-4o as the evaluator. The knowledge graph employed by the KEO method is constructed using Phi-4-mini with varying numbers of aviation records.}
\label{tab:senseQA_kg-phi_eval-gpt}
\renewcommand{\arraystretch}{1.8} 
\begin{adjustbox}{max width=\linewidth}
{\LARGE
\begin{tabular}{l *{5}{ccc}}
\toprule
\diagbox{Model}{KG type} 
& \multicolumn{3}{c}{\textbf{Phi-4-mini-100}} 
& \multicolumn{3}{c}{\textbf{Phi-4-mini-200}} 
& \multicolumn{3}{c}{\textbf{Phi-4-mini-300}} 
& \multicolumn{3}{c}{\textbf{Phi-4-mini-400}} 
& \multicolumn{3}{c}{\textbf{Phi-4-mini-500}} \\
\cmidrule(lr){2-4} \cmidrule(lr){5-7} \cmidrule(lr){8-10} \cmidrule(lr){11-13} \cmidrule(lr){14-16}

& \textbf{TC} & \textbf{VN} & \textbf{KG}
& \textbf{TC} & \textbf{VN} & \textbf{KG}
& \textbf{TC} & \textbf{VN} & \textbf{KG}
& \textbf{TC} & \textbf{VN} & \textbf{KG}
& \textbf{TC} & \textbf{VN} & \textbf{KG} \\
\midrule

\textbf{gemma-3-it}  & $4.05 \pm 0.56$ & $3.68 \pm 0.62$ & $4.31 \pm 0.29$ & $4.11 \pm 0.44$ & $3.72 \pm 0.53$ & \boldmath $4.33 \pm 0.29$ \unboldmath & $4.15 \pm 0.41$ & $3.71 \pm 0.54$ & $4.29 \pm 0.30$ & $4.11 \pm 0.52$ & $3.70 \pm 0.50$ & $4.27 \pm 0.37$ & $4.14 \pm 0.42$ & $3.74 \pm 0.50$ & $4.26 \pm 0.35$
\\
\textbf{phi-4}  & $3.90 \pm 0.47$ & $3.71 \pm 0.51$ & $4.01 \pm 0.42$ & $3.86 \pm 0.53$ & $3.64 \pm 0.54$ & $4.06 \pm 0.42$ & $3.86 \pm 0.51$ & $3.68 \pm 0.55$ & \boldmath $4.13 \pm 0.40$ \unboldmath & $3.89 \pm 0.50$ & $3.68 \pm 0.53$ & $4.09 \pm 0.41$ & $3.95 \pm 0.47$ & $3.69 \pm 0.49$ & $4.01 \pm 0.42$
\\
\textbf{mistral-nemo-it}  & $3.82 \pm 0.57$ & $3.46 \pm 0.54$ & $3.81 \pm 0.50$ & $3.77 \pm 0.60$ & $3.43 \pm 0.72$ & \boldmath $3.85 \pm 0.49$ \unboldmath & $3.78 \pm 0.59$ & $3.47 \pm 0.58$ & $3.80 \pm 0.52$ & $3.82 \pm 0.56$ & $3.46 \pm 0.54$ & $3.76 \pm 0.50$ & $3.74 \pm 0.67$ & $3.49 \pm 0.60$ & $3.75 \pm 0.46$
\\
\bottomrule
\end{tabular}
}
\end{adjustbox}
\end{table*}

\begin{table*}[htbp]
\centering
\caption{Evaluation results on 83 global sensemaking questions using Llama-3.3-70B-Instruct as the evaluator. The knowledge graph employed by the KEO method is constructed using GPT-4o with varying numbers of aviation records.}
\label{tab:senseQA_kg-gpt_eval-llama}
\renewcommand{\arraystretch}{1.8} 
\begin{adjustbox}{max width=\linewidth}
{\LARGE
\begin{tabular}{l *{5}{ccc}}
\toprule
\diagbox{Model}{KG type} 
& \multicolumn{3}{c}{\textbf{GPT-4o-100}} 
& \multicolumn{3}{c}{\textbf{GPT-4o-200}} 
& \multicolumn{3}{c}{\textbf{GPT-4o-300}} 
& \multicolumn{3}{c}{\textbf{GPT-4o-400}} 
& \multicolumn{3}{c}{\textbf{GPT-4o-500}} \\
\cmidrule(lr){2-4} \cmidrule(lr){5-7} \cmidrule(lr){8-10} \cmidrule(lr){11-13} \cmidrule(lr){14-16}
& \textbf{TC} & \textbf{VN} & \textbf{KG}
& \textbf{TC} & \textbf{VN} & \textbf{KG}
& \textbf{TC} & \textbf{VN} & \textbf{KG}
& \textbf{TC} & \textbf{VN} & \textbf{KG}
& \textbf{TC} & \textbf{VN} & \textbf{KG} \\
\midrule

\textbf{gemma-3-it}  & $4.47 \pm 0.45$ & $4.20 \pm 0.62$ & $4.88 \pm 0.25$ & $4.52 \pm 0.28$ & $4.31 \pm 0.23$ & \boldmath $4.93 \pm 0.20$ \unboldmath & $4.49 \pm 0.27$ & $4.32 \pm 0.21$ & $4.90 \pm 0.22$ & $4.47 \pm 0.23$ & $4.30 \pm 0.28$ & $4.91 \pm 0.22$ & $4.49 \pm 0.23$ & $4.24 \pm 0.47$ & $4.93 \pm 0.21$
\\
\textbf{phi-4}  & $4.34 \pm 0.13$ & $4.31 \pm 0.21$ & $4.48 \pm 0.21$ & $4.32 \pm 0.21$ & $4.29 \pm 0.25$ & $4.47 \pm 0.20$ & $4.29 \pm 0.42$ & $4.30 \pm 0.23$ & \boldmath $4.50 \pm 0.23$ \unboldmath & $4.29 \pm 0.40$ & $4.28 \pm 0.32$ & $4.42 \pm 0.42$ & $4.32 \pm 0.20$ & $4.29 \pm 0.25$ & $4.45 \pm 0.45$
\\
\textbf{mistral-nemo-it} & $4.31 \pm 0.26$ & $4.07 \pm 0.61$ & $4.34 \pm 0.45$ & $4.28 \pm 0.31$ & $4.13 \pm 0.43$ & \boldmath $4.42 \pm 0.17$ \unboldmath & $4.20 \pm 0.39$ & $4.09 \pm 0.49$ & $4.40 \pm 0.13$ & $4.28 \pm 0.28$ & $4.12 \pm 0.50$ & $4.37 \pm 0.16$ & $4.27 \pm 0.33$ & $4.11 \pm 0.49$ & $4.39 \pm 0.22$
\\
\bottomrule
\end{tabular}
}
\end{adjustbox}
\end{table*}

\begin{table*}[htbp]
\centering
\caption{Evaluation results on 83 global sensemaking questions using Llama-3.3-70B-Instruct as the evaluator. The knowledge graph employed by the KEO method is constructed using Phi-4-mini with varying numbers of aviation records.}
\label{tab:senseQA_kg-phi_eval-llama}
\renewcommand{\arraystretch}{1.8} 
\begin{adjustbox}{max width=\linewidth}
{\LARGE
\begin{tabular}{l *{5}{ccc}}
\toprule
\diagbox{Model}{KG type} 

& \multicolumn{3}{c}{\textbf{Phi-4-mini-100}} 
& \multicolumn{3}{c}{\textbf{Phi-4-mini-200}} 
& \multicolumn{3}{c}{\textbf{Phi-4-mini-300}} 
& \multicolumn{3}{c}{\textbf{Phi-4-mini-400}} 
& \multicolumn{3}{c}{\textbf{Phi-4-mini-500}} \\
\cmidrule(lr){2-4} \cmidrule(lr){5-7} \cmidrule(lr){8-10} \cmidrule(lr){11-13} \cmidrule(lr){14-16}

& \textbf{TC} & \textbf{VN} & \textbf{KG}
& \textbf{TC} & \textbf{VN} & \textbf{KG}
& \textbf{TC} & \textbf{VN} & \textbf{KG}
& \textbf{TC} & \textbf{VN} & \textbf{KG}
& \textbf{TC} & \textbf{VN} & \textbf{KG} \\
\midrule

\textbf{gemma-3-it}  & $4.45 \pm 0.44$ & $4.25 \pm 0.46$ & $4.85 \pm 0.27$ & $4.48 \pm 0.22$ & $4.33 \pm 0.23$ & $4.81 \pm 0.28$ & $4.50 \pm 0.26$ & $4.28 \pm 0.28$ & \boldmath $4.88 \pm 0.24$ \unboldmath & $4.43 \pm 0.44$ & $4.27 \pm 0.30$ & $4.84 \pm 0.26$ & $4.50 \pm 0.26$ & $4.31 \pm 0.28$ & $4.84 \pm 0.27$
\\
\textbf{phi-4}  & $4.33 \pm 0.21$ & $4.32 \pm 0.21$ & $4.47 \pm 0.20$ & $4.33 \pm 0.17$ & $4.27 \pm 0.31$ & $4.48 \pm 0.19$ & $4.33 \pm 0.18$ & $4.30 \pm 0.24$ & \boldmath $4.50 \pm 0.22$ \unboldmath & $4.32 \pm 0.21$ & $4.34 \pm 0.19$ & $4.46 \pm 0.18$ & $4.33 \pm 0.18$ & $4.32 \pm 0.19$ & $4.45 \pm 0.17$
\\
\textbf{mistral-nemo-it} & $4.26 \pm 0.30$ & $4.11 \pm 0.44$ & $4.39 \pm 0.12$ & $4.31 \pm 0.27$ & $4.00 \pm 0.67$ & \boldmath $4.42 \pm 0.14$ \unboldmath & $4.32 \pm 0.28$ & $4.11 \pm 0.48$ & $4.39 \pm 0.19$ & $4.26 \pm 0.34$ & $4.15 \pm 0.44$ & $4.37 \pm 0.19$ & $4.26 \pm 0.38$ & $4.10 \pm 0.52$ & $4.41 \pm 0.18$

\\
\bottomrule
\end{tabular}
}
\end{adjustbox}
\end{table*}

\begin{table*}[htbp]
\centering
\caption{Evaluation results on 50 generalizable knowledge-to-action questions using GPT-4o as the evaluator. The knowledge graph employed by the KEO method is constructed using Phi-4-mini with varying numbers of aviation records.}
\label{tab:actionQA_kg-phi_eval-gpt}
\renewcommand{\arraystretch}{1.8} 
\begin{adjustbox}{max width=\linewidth}
{\LARGE
\begin{tabular}{l *{5}{ccc}}
\toprule
\diagbox{Model}{KG type} 

& \multicolumn{3}{c}{\textbf{Phi-4-mini-100}} 
& \multicolumn{3}{c}{\textbf{Phi-4-mini-200}} 
& \multicolumn{3}{c}{\textbf{Phi-4-mini-300}} 
& \multicolumn{3}{c}{\textbf{Phi-4-mini-400}} 
& \multicolumn{3}{c}{\textbf{Phi-4-mini-500}} \\
\cmidrule(lr){2-4} \cmidrule(lr){5-7} \cmidrule(lr){8-10} \cmidrule(lr){11-13} \cmidrule(lr){14-16}

& \textbf{TC} & \textbf{VN} & \textbf{KG}
& \textbf{TC} & \textbf{VN} & \textbf{KG}
& \textbf{TC} & \textbf{VN} & \textbf{KG}
& \textbf{TC} & \textbf{VN} & \textbf{KG}
& \textbf{TC} & \textbf{VN} & \textbf{KG} \\
\midrule

\textbf{gemma-3-it}  & \boldmath $3.89 \pm 0.66$ \unboldmath & $3.84 \pm 0.63$ & $3.77 \pm 0.69$ & $3.82 \pm 0.68$ & $3.83 \pm 0.64$ & $3.77 \pm 0.72$ & $3.86 \pm 0.65$ & $3.77 \pm 0.67$ & $3.72 \pm 0.74$ & $3.86 \pm 0.63$ & $3.78 \pm 0.68$ & $3.78 \pm 0.64$ & $3.85 \pm 0.66$ & $3.84 \pm 0.65$ & $3.81 \pm 0.69$
\\
\textbf{phi-4}  & $4.14 \pm 0.45$ & $4.00 \pm 0.55$ & $3.97 \pm 0.53$ & \boldmath $4.21 \pm 0.45$ \unboldmath & $4.02 \pm 0.59$ & $3.90 \pm 0.53$ & $4.14 \pm 0.47$ & $4.04 \pm 0.52$ & $3.86 \pm 0.67$ & $4.14 \pm 0.57$ & $4.04 \pm 0.50$ & $3.91 \pm 0.60$ & $4.15 \pm 0.45$ & $4.05 \pm 0.48$ & $3.92 \pm 0.54$
\\
\textbf{mistral-nemo-it}  & $3.71 \pm 0.67$ & \boldmath $3.75 \pm 0.79$ \unboldmath & $3.70 \pm 0.78$ & $3.68 \pm 0.76$ & $3.73 \pm 0.79$ & $3.68 \pm 0.86$ & $3.65 \pm 0.74$ & $3.62 \pm 0.85$ & $3.51 \pm 0.89$ & $3.70 \pm 0.69$ & $3.66 \pm 0.82$ & $3.58 \pm 0.87$ & $3.66 \pm 0.73$ & $3.70 \pm 0.74$ & $3.49 \pm 0.91$
\\
\bottomrule
\end{tabular}
}
\end{adjustbox}
\end{table*}

\begin{table*}[htbp]
\centering
\caption{Evaluation results on 50 generalizable knowledge-to-action questions using Llama-3.3-70B-Instruct as the evaluator. The knowledge graph employed by the KEO method is constructed using GPT-4o with varying numbers of aviation records.}
\label{tab:actionQA_kg-gpt_eval-llama}
\renewcommand{\arraystretch}{1.8} 
\begin{adjustbox}{max width=\linewidth}
{\LARGE
\begin{tabular}{l *{5}{ccc}}
\toprule
\diagbox{Model}{KG type} 

& \multicolumn{3}{c}{\textbf{GPT-4o-100}} 
& \multicolumn{3}{c}{\textbf{GPT-4o-200}} 
& \multicolumn{3}{c}{\textbf{GPT-4o-300}} 
& \multicolumn{3}{c}{\textbf{GPT-4o-400}} 
& \multicolumn{3}{c}{\textbf{GPT-4o-500}} \\
\cmidrule(lr){2-4} \cmidrule(lr){5-7} \cmidrule(lr){8-10} \cmidrule(lr){11-13} \cmidrule(lr){14-16} 

& \textbf{TC} & \textbf{VN} & \textbf{KG}
& \textbf{TC} & \textbf{VN} & \textbf{KG}
& \textbf{TC} & \textbf{VN} & \textbf{KG}
& \textbf{TC} & \textbf{VN} & \textbf{KG}
& \textbf{TC} & \textbf{VN} & \textbf{KG} \\
\midrule

\textbf{gemma-3-it} & $3.98 \pm 0.60$ & $3.96 \pm 0.61$ & $3.77 \pm 0.78$ & $4.00 \pm 0.59$ & $3.96 \pm 0.62$ & $3.85 \pm 0.68$ & \boldmath $4.00 \pm 0.54$ \unboldmath & $3.97 \pm 0.61$ & $3.81 \pm 0.74$ & $3.99 \pm 0.60$ & $3.87 \pm 0.71$ & $3.87 \pm 0.69$ & $4.00 \pm 0.60$ & $3.94 \pm 0.59$ & $3.78 \pm 0.74$
\\
\textbf{phi-4} & $4.32 \pm 0.37$ & $4.27 \pm 0.39$ & $4.19 \pm 0.41$ & $4.33 \pm 0.36$ & $4.24 \pm 0.43$ & $4.17 \pm 0.44$ & \boldmath $4.36 \pm 0.33$ \unboldmath & $4.24 \pm 0.43$ & $4.23 \pm 0.46$ & $4.32 \pm 0.36$ & $4.29 \pm 0.36$ & $4.15 \pm 0.47$ & $4.32 \pm 0.33$ & $4.26 \pm 0.41$ & $4.14 \pm 0.47$
\\
\textbf{mistral-nemo-it} & \boldmath $4.03 \pm 0.53$ \unboldmath & $3.87 \pm 0.73$ & $3.75 \pm 0.77$ & $3.96 \pm 0.63$ & $3.81 \pm 0.78$ & $3.84 \pm 0.68$ & $3.92 \pm 0.62$ & $3.80 \pm 0.80$ & $3.78 \pm 0.79$ & $4.00 \pm 0.60$ & $3.76 \pm 0.80$ & $3.74 \pm 0.78$ & $3.83 \pm 0.71$ & $3.82 \pm 0.74$ & $3.82 \pm 0.76$
\\
\bottomrule
\end{tabular}
}
\end{adjustbox}
\end{table*}

\begin{table*}[htbp]
\centering
\caption{Evaluation results on 50 generalizable knowledge-to-action questions using Llama-3.3-70B-Instruct as the evaluator. The knowledge graph employed by the KEO method is constructed using Phi-4-mini with varying numbers of aviation records.}
\label{tab:actionQA_kg-phi_eval-llama}
\renewcommand{\arraystretch}{1.8} 
\begin{adjustbox}{max width=\linewidth}
{\LARGE
\begin{tabular}{l *{5}{ccc}}
\toprule
\diagbox{Model}{KG type} 
& \multicolumn{3}{c}{\textbf{Phi-4-mini-100}} 
& \multicolumn{3}{c}{\textbf{Phi-4-mini-200}} 
& \multicolumn{3}{c}{\textbf{Phi-4-mini-300}} 
& \multicolumn{3}{c}{\textbf{Phi-4-mini-400}} 
& \multicolumn{3}{c}{\textbf{Phi-4-mini-500}} \\
\cmidrule(lr){2-4} \cmidrule(lr){5-7} \cmidrule(lr){8-10} \cmidrule(lr){11-13} \cmidrule(lr){14-16} 

& \textbf{TC} & \textbf{VN} & \textbf{KG}
& \textbf{TC} & \textbf{VN} & \textbf{KG}
& \textbf{TC} & \textbf{VN} & \textbf{KG}
& \textbf{TC} & \textbf{VN} & \textbf{KG}
& \textbf{TC} & \textbf{VN} & \textbf{KG} \\
\midrule

\textbf{gemma-3-it}  & \boldmath $4.00 \pm 0.59$ \unboldmath & $3.93 \pm 0.62$ & $3.85 \pm 0.71$ & \boldmath $4.00 \pm 0.59$ \unboldmath & $3.97 \pm 0.60$ & $3.83 \pm 0.72$ & $3.99 \pm 0.59$ & $3.94 \pm 0.61$ & $3.80 \pm 0.76$ & $4.00 \pm 0.62$ & $3.95 \pm 0.62$ & $3.86 \pm 0.64$ & $4.00 \pm 0.62$ & $3.95 \pm 0.62$ & $3.85 \pm 0.69$
\\
\textbf{phi-4}  & \boldmath $4.35 \pm 0.31$ \unboldmath & $4.20 \pm 0.45$ & $4.18 \pm 0.46$ & $4.33 \pm 0.37$ & $4.26 \pm 0.45$ & $4.19 \pm 0.43$ & $4.32 \pm 0.37$ & $4.22 \pm 0.43$ & $4.11 \pm 0.62$ & $4.26 \pm 0.58$ & $4.28 \pm 0.39$ & $4.17 \pm 0.50$ & $4.32 \pm 0.36$ & $4.27 \pm 0.44$ & $4.15 \pm 0.46$
\\
\textbf{mistral-nemo-it}  & \boldmath $3.92 \pm 0.56$ \unboldmath & $3.88 \pm 0.71$ & $3.87 \pm 0.67$ & $3.92 \pm 0.68$ & $3.83 \pm 0.76$ & $3.80 \pm 0.76$ & $3.88 \pm 0.62$ & $3.76 \pm 0.80$ & $3.64 \pm 0.86$ & $3.92 \pm 0.60$ & $3.73 \pm 0.79$ & $3.70 \pm 0.78$ & $3.83 \pm 0.65$ & $3.76 \pm 0.75$ & $3.61 \pm 0.85$
\\
\bottomrule
\end{tabular}
}
\end{adjustbox}
\end{table*}

\section{KG Creation Prompt}
\label{sec:kg_creation_prompt}
Here we provide the prompt used to extract knowledge graph triplets from aviation maintenance text, enabling structured knowledge representation for the KEO framework.

{\small
\setlength{\parindent}{0pt}
\textbf{Role---}  
You are extracting knowledge graph triplets from aviation maintenance text.

\textbf{Task---}  
Extract informative triplets directly from the text following the examples. Format each triplet as: \texttt{<entity1, relation, entity2>}. Do not add any extra words, line breaks, or explanatory notes. Focus on extracting factual relationships from the text.

\textbf{Goal---}  
Generate structured entity-relation-entity triplets that capture factual relationships from aviation maintenance records while maintaining consistency across the knowledge graph.

\textbf{Guidelines---}  
Use only these relation types: \texttt{OWNED BY}, \texttt{INSTANCE OF}, \texttt{FOLLOWED BY}, \texttt{HAS CAUSE}, \texttt{FOLLOWS}, \texttt{EVENT DISTANCE}, \texttt{HAS EFFECT}, \texttt{LOCATION}, \texttt{USED BY}, \texttt{INFLUENCED BY}, \texttt{TIME PERIOD}, \texttt{PART OF}, \texttt{MAINTAINED BY}, \texttt{DESIGNED BY}. When extracting triplets, prefer to use existing nodes from the knowledge graph if possible, rather than inventing new entity mentions.

\textbf{Style---}  
Extract triplets in the format \texttt{<entity1, relation, entity2>}. Each triplet should appear on a new line with no numbering or bullets. Focus on factual relationships that can be directly inferred from the text.

\textbf{Example---}  
\texttt{TEXT: THE WRIGHT BROTHERS DESIGNED THE FIRST SUCCESSFUL AIRPLANE IN 1903 IN KITTY HAWK.}\\
\texttt{Triplets:}\\
\texttt{<FIRST SUCCESSFUL AIRPLANE, DESIGNED BY, WRIGHT BROTHERS>}\\
\texttt{<FIRST SUCCESSFUL AIRPLANE, TIME PERIOD, 1903>}\\
\texttt{<FIRST SUCCESSFUL AIRPLANE, LOCATION, KITTY HAWK>}

\textbf{Target Text:} \texttt{\{text\}}\\
\textbf{Triplets:}
\par}
\section{Global Sensemaking Question Prompts}
\label{sec:sensemaking_QA_prompt}
Here we provide the prompts used to generate global sensemaking questions for the QA benchmark, covering three subtypes: comprehensive, context-specific, and category-specific.

\subsection{Comprehensive Question Prompt}
{\small
\setlength{\parindent}{0pt}
\textbf{Role---}  
You are an expert in aviation safety and maintenance analysis.

\textbf{Task---}  
Generate high-level, global sensemaking questions based on a summary of an aviation maintenance dataset. These questions should require a comprehensive understanding of patterns and relationships that span the entire dataset.

\textbf{Goal---}  
Generate exactly \texttt{\{num\_questions\}} questions that encourage holistic insight into overarching themes, systemic issues, and strategic implications present in the dataset. The dataset summary is provided below: \texttt{\{data\_summary\}}.

\textbf{Guidelines---}  
The questions must: (1) require a holistic understanding of the dataset, not just individual records; (2) focus on patterns, trends, causal relationships, and systemic insights; (3) enable strategic reasoning or decision-making at an organizational level; and (4) avoid questions that can be answered through simple statistics or localized analysis.

\textbf{Style---}  
Generate exactly \texttt{\{num\_questions\}} questions. Each question should appear on a new line, with no numbering or bullets. Use formulations such as: “What are the...”, “How do...”, “Which factors...”, “What patterns...”.

\textbf{Examples---}  
What are the most common systemic maintenance challenges reflected across the dataset?  
How do maintenance trends vary by aircraft type, and what are the implications?  
Which recurring issues suggest gaps in standard operating procedures?  
What are the long-term safety patterns indicated by the data?
\par}

\subsection{Context-Specific Question Prompt}

{\small
\setlength{\parindent}{0pt}
\textbf{Role---}  
You are an expert in aviation safety and maintenance analysis.

\textbf{Task---}  
Generate context-specific sensemaking questions based on representative aviation maintenance records. These questions should help uncover causes, patterns, and interactions relevant to the particular maintenance context.

\textbf{Goal---}  
Given representative maintenance records from the \texttt{\{context\_type\}} context: \texttt{\{sample\_records\}}, generate exactly \texttt{\{num\_questions\}} questions that promote analysis across multiple records and inform actionable understanding.

\textbf{Guidelines---}  
The questions should:  
(1) Focus on understanding \textit{why} failures occur.  
(2) Identify \textit{what} can be done to prevent similar issues.  
(3) Explore \textit{how} different factors interact.  
(4) Reveal patterns across similar incidents.  
(5) Support decision-making for maintenance and safety.  

\textbf{Style---}  
Generate exactly \texttt{\{num\_questions\}} questions. Each question should appear on a new line, with no numbering or bullets. Focus on questions that require synthesis of multiple incidents or factors, rather than simple summaries or isolated examples.
\par}

\subsection{Category-Specific Question Prompt}

{\small
\setlength{\parindent}{0pt}
\textbf{Role---}  
You are an expert in aviation safety and maintenance analysis.

\textbf{Task---}  
Generate high-quality sensemaking questions tailored to a specific analytical category in aviation maintenance.

\textbf{Goal---}  
Generate exactly \texttt{\{num\_questions\}} questions that align with the category: \texttt{\{category\}}. These questions should reflect deep insight into aviation maintenance practices and support strategic reasoning.

\textbf{Category---}  
\texttt{\{config['description']\}}

\textbf{Context---}  
\texttt{\{context\_prompt\}}

\textbf{Guidelines---}  
The questions should:  
(1) Require synthesis across multiple data points to answer.  
(2) Focus on actionable insights for aviation safety.  
(3) Be specific to the aviation maintenance domain.  
(4) Support strategic decision-making.  
(5) Reveal patterns and relationships in the data.

\textbf{Style---}  
Use the following starter patterns as inspiration, but generate varied, comprehensive questions: \texttt{\{config['template\_starters']\}}  
Generate exactly \texttt{\{num\_questions\}} questions. Each question should appear on a new line, with no numbering or bullets.
\par}

\section{Task-Specific Evaluation Metrics}
\label{sec:evalmetric}
\textbf{Global Sensemaking Questions.}  
Since these questions have no predefined gold answers, we use an LLM-based evaluator for both absolute and comparative scoring. For absolute scoring, the evaluator rates each answer on a 1–5 scale according to task-specific criteria, providing detailed explanations. For comparative scoring, pairwise comparisons are conducted between answers from different methods (Vanilla LLM, Text-chunk RAG, and KEO). The evaluator selects the better answer for each pair, and we report the win rate of each method across the full set of comparisons.

The evaluation is guided by these criteria:
\begin{itemize}[leftmargin=10pt, itemsep=0pt, parsep=0pt]
    \item \textbf{Global Perspective:} Does the answer reflect dataset-wide insights?
    \item \textbf{Theme Identification:} Are key recurring patterns or topics clearly identified?
    \item \textbf{Synthesis Quality:} Does the response integrate information across different records?
    \item \textbf{Strategic Value:} Are the insights useful for high-level decision-making?
    \item \textbf{Pattern Recognition:} Are underlying trends or systemic relationships revealed?
\end{itemize}

\textbf{Knowledge-to-Action Questions.}  
We apply both automatic and LLM-based evaluation. For automatic evaluation, we compute ROUGE-F1 scores between predicted and gold-standard actions. For human-aligned evaluation, we prompt an LLM with the question, predicted answer, and gold-standard answer, and ask it to rate the response using the following criteria (1–5 scale):
\begin{itemize}[leftmargin=10pt, itemsep=0pt, parsep=0pt]
    \item \textbf{Correctness:} Is the predicted action factually accurate?
    \item \textbf{Completeness:} Does it cover all necessary action steps?
    \item \textbf{Practicality:} Is the recommended action feasible in real-world maintenance?
    \item \textbf{Safety:} Does the response preserve or enhance operational safety?
    \item \textbf{Clarity:} Is the action clearly and precisely articulated?
\end{itemize}

\section{LLM Evaluator Prompts}
\label{sec:evaluator_prompt}
Here we provide the prompts used by the LLM evaluator to assess both global sensemaking and knowledge-to-action questions.

\subsection{Global Sensemaking Evaluation Prompt}
{\small
\setlength{\parindent}{0pt}
\textbf{Role---}  
You are evaluating an LLM-generated answer to a \textbf{global sensemaking} question in the domain of aviation maintenance.

\textbf{Input---}  
Question: \{\texttt{question}\};
LLM Answer: \{\texttt{answer}\} 

\textbf{Definition---}  
Global sensemaking questions require synthesis across entire datasets to identify overarching themes, systemic patterns, and strategic insights.

\textbf{Evaluation Criteria---}  
Rate the answer on a 1–5 scale for each of the following:
\vspace{-0.1in}
\begin{itemize}[leftmargin=10pt, itemsep=0pt, parsep=0pt]
  \item \textbf{Global Perspective:} Does the answer demonstrate understanding of dataset-wide patterns?
  \item \textbf{Theme Identification:} Are major themes and patterns clearly identified?
  \item \textbf{Synthesis Quality:} How well does it synthesize information across multiple sources?
  \item \textbf{Strategic Value:} Does it provide insights useful for high-level decision making?
  \item \textbf{Pattern Recognition:} Are complex relationships and dependencies identified?
\end{itemize}

\textbf{Output Format---}  

\texttt{Global Perspective: [score] - [explanation]}\\
\texttt{Theme Identification: [score] - [explanation]}\\
\texttt{Synthesis Quality: [score] - [explanation]}\\
\texttt{Strategic Value: [score] - [explanation]}\\
\texttt{Pattern Recognition: [score] - [explanation]}\\
\texttt{Global Sensemaking Assessment: [overall score]}
\par}

\subsection{Knowledge-to-Action Evaluation Prompt}
{\small
\setlength{\parindent}{0pt}
\textbf{Role---}  
You are evaluating a predicted answer to a maintenance-action question, assessing its correctness and safety based on the gold-standard response.

\textbf{Input---}  
\texttt{Question: \{question\}}  
\texttt{Ground Truth Answer: \{ground\_truth\}}  
\texttt{Predicted Answer: \{predicted\}}

\textbf{Evaluation Criteria---}  
Rate on a 1–5 scale:
\vspace{-0.1in}
\begin{itemize}[leftmargin=10pt, itemsep=0pt, parsep=0pt]
  \item \textbf{Correctness:} How factually accurate is the predicted answer?
  \item \textbf{Completeness:} Does it cover all necessary action steps?
  \item \textbf{Practicality:} Are the actions feasible and implementable?
  \item \textbf{Safety:} Would the actions maintain or improve operational safety?
  \item \textbf{Clarity:} Is the answer easy to understand and follow?
\end{itemize}
\textbf{Output Format---}

\texttt{Correctness: [score] - [explanation]}\\
\texttt{Completeness: [score] - [explanation]}\\
\texttt{Practicality: [score] - [explanation]}\\
\texttt{Safety: [score] - [explanation]}\\
\texttt{Clarity: [score] - [explanation]}\\
\texttt{Overall Score: [average score] - [summary of quality]}
\par}

\section{Relation Extraction}
\subsection{Lack of a Relation Extraction Gold Standard}
\label{sec:lack_REGS}
The OMIn dataset has previously been processed through a knowledge extraction pipeline covering NER, CR, NEL, and RE~\cite{mealey2025trusted}. A random subset of 100 records was human-annotated to provide gold standards for NER, CR, and NEL, with the same subset manually reviewed for RE. 

The original FAA dataset contains sentences such as: ``After takeoff, engine quit. Wing fuel tank sumps were not drained during preflight because they were frozen.''~\cite{faa_accident_incident_dataset}. Although prior work demonstrated that a KG could in principle be derived from such records~\cite{mealey2025trusted}, the construction of a gold standard for RE was deferred. This omission stemmed from the fact that each RE tool under evaluation employed a distinct relation schema, making a unified global GS impractical. However, since our KG–RAG pipeline requires structured relations, we address this gap by creating a dedicated RE gold standard.

\begin{table}[t]
\centering
\caption{Distribution of relation types in the gold standard for RE.}
\label{tab:relation_counts}
\renewcommand{\arraystretch}{1.05} 
\begin{adjustbox}{max width=0.9\columnwidth} 
\begin{tabular}{lcc}
\toprule
\textbf{Relation} & \textbf{Strict Count} & \textbf{Loose Count} \\
\midrule
OWNED BY       & 1  & 2  \\
INSTANCE OF    & 5  & 9  \\
FOLLOWED BY    & 0  & 20 \\
HAS CAUSE      & 13 & 93 \\
FOLLOWS        & 1  & 8  \\
EVENT DISTANCE & 0  & 1  \\
HAS EFFECT     & 13 & 94 \\
LOCATION       & 12 & 23 \\
USED BY        & 4  & 5  \\
INFLUENCED BY  & 0  & 1  \\
TIME PERIOD    & 6  & 29 \\
PART OF        & 21 & 30 \\
MAINTAINED BY  & 8  & 9  \\
DESIGNED BY    & 0  & 1  \\
\bottomrule
\end{tabular}
\end{adjustbox}
\end{table}

\subsection{Creation of a Relation Extraction Gold Standard}
\label{sec:REGS}
In their evaluation of RE tools, prior studies compared four different systems, each adopting its own set of relations~\cite{mealey2025trusted}. To establish consistent gold standards, we defined a relation schema that combines the REBEL subset of Wikidata relations~\cite{huguet-cabot-navigli-2021-rebel-relation} with additional domain-specific relations relevant to aviation maintenance. Specifically, we included relations such as \texttt{MAINTAINED BY} and \texttt{DESIGNED BY} to capture critical maintenance-specific dependencies (Table~\ref{tab:relation_counts}). 

This schema was selected because the REBEL tool outperformed others in prior experiments on the OMIn dataset, identifying 220 relations and providing a comprehensive basis for KG construction. Using this relation set, we created 100 gold-standard annotations through a standard human annotation process.

\textbf{RE Strict Gold Standard} - In order to create a relationship triple (head, relationship, tail), the head and tail entities must already exist as nodes from the NER or CR processes in order for the relationship to be added.

\textbf{RE Loose Gold Standard} - If a head or tail node does not already exist, a new node will be created when the relationship is added.

The loose count was introduced to mitigate the limitations of generating the KG via a single, complex prompt per record.

This metric served as a relaxed evaluation by only requiring that the LLM to extract a plausible relation type from the correct record ID. This less strict measure allowed for a fairer assessment of the LLM's core ability to identify relevant relational concepts within the complex input text, despite the difficulty posed by the single-prompt extraction methodology, as noted in Appendix \ref{sec:kg_creation_prompt}.

\section{Study Scope and Constraints}
\label{sec:scope}
KEO framework's evaluation is  constrained to ensure the integrity of the knowledge-transfer assessment and the viability of real-world application. The research limits the evaluation to locally deployable LLMs, such as Gemma-3, Phi-4, and Mistral-Nemo, as this proves the framework's suitability for secure, in-house deployment in safety-critical domains where large, proprietary cloud APIs are unsuitable. Stronger models, such as GPT-4o and Llama-3.3, are employed solely as judges for a robust, objective evaluation. To rigorously test the transferability of structured knowledge, the target answer corpus, MaintNet records \cite{akhbardeh2020maintnet}, were excluded from the retrieval corpus; instead, both RAG and KEO are strictly restricted to the OMIn dataset corpus only to force the systems to answer by leveraging transferred maintenance knowledge successfully. This domain focus highlights the inherent challenge of working in this specialized field, as there is a significant lack of data from the U.S. military and defense sectors in the public domain, with the few available records. Others include NASA's Prognostics Center of Excellence (PCOE) \cite{nasa_pcoe}, which offers no free-response natural language text, and NASA's Aviation Safety Reporting System (ASRS) \cite{nasa_asrs_dataset}, which provides only a limited selection of 50 records regarding 30 topics.

\end{document}